\documentclass[10pt,twocolumn,letterpaper]{article}

\usepackage[]{cvpr}      

\usepackage{tikz}
\usetikzlibrary{positioning, arrows.meta, calc}

\definecolor{cvprblue}{rgb}{0.21,0.49,0.74}
\usepackage[pagebackref,breaklinks,colorlinks,allcolors=cvprblue]{hyperref}

\def\paperID{*****} 
\def\confName{CVPR}
\def\confYear{2026}

\title{Conflict-Aware Multimodal Fusion for Ambivalence and Hesitancy Recognition}

\author{
Salah Eddine Bekhouche\\
University of the Basque Country UPV/EHU, San Sebastian, Spain\\
{\tt\small bekhouchesalah@gmail.com}
\and
Hichem Telli\\
Laboratory of LESIA, University of Biskra, Algeria\\
{\tt\small tellihicham@univ-biskra.dz}
\and
Azeddine Benlamoudi\\
Lab. de Génie Electrique (LAGE), University Kasdi Merbah Ouargla, Ouargla, Algeria\\
{\tt\small benlamoudi.azeddine@univ-ouargla.dz}
\and
Salah Eddine Herrouz\\
Lab. de Génie Electrique (LAGE), University Kasdi Merbah Ouargla, Ouargla, Algeria\\
{\tt\small herrouz.salaheddine@univ-ouargla.dz }
\and
Abdelmalik Taleb-Ahmed\\
Institute of Electronics, Microelectronics and Nanotechnology (IEMN), \\
Polytechnic University of Hauts-de-France, University of Lille, Valenciennes, France\\
{\tt\small abdelmalik.taleb-ahmed@uphf.fr }
\and
Abdenour Hadid\\
Sorbonne Center for Artificial Intelligence, Sorbonne University Abu Dhabi, UAE\\
{\tt\small abdenour.hadid@sorbonne.ae }
}

\begin{document}
\maketitle
\begin{abstract}
Ambivalence and hesitancy (A/H) are subtle affective states where a person
shows conflicting signals through different channels---saying one thing while
their face or voice tells another story.
Recognising these states automatically is valuable in clinical settings, but it
is hard for machines because the key evidence lives in the \emph{disagreements}
between what is said, how it sounds, and what the face shows.

We present \textbf{ConflictAwareAH}, a multimodal framework built for this
problem.
Three pre-trained encoders extract video, audio, and text representations.
Pairwise conflict features---element-wise absolute differences between modality
embeddings---serve as \emph{bidirectional} cues: large cross-modal differences
flag A/H, while small differences confirm behavioural consistency and anchor
the negative class.
This conflict-aware design addresses a key limitation of text-dominant
approaches, which tend to over-detect A/H (high F1-AH) while struggling to
confirm its absence: our multimodal model improves F1-NoAH by +4.6 points
over text alone and halves the class-performance gap.
A complementary \emph{text-guided late fusion} strategy blends a text-only
auxiliary head with the full model at inference, adding +4.1 Macro F1.

On the BAH dataset from the ABAW10 Ambivalence/Hesitancy Challenge, our
method reaches \textbf{0.694 Macro F1} on the labelled test split and
\textbf{0.715} on the private leaderboard, outperforming published
multimodal baselines by over 10 points---all on a single GPU in under 25
minutes of training.
\end{abstract}

\section{Introduction}

Ambivalence is when someone holds conflicting feelings about the same thing at
the same time. Hesitancy is closely related: it is the reluctance or the delay
before committing to a decision.
Both states come up frequently in clinical psychology, behavioural economics,
and human--computer interaction, yet they remain difficult to detect
automatically.

Why is this clinically important?
In medical interviews, a patient may agree to a treatment plan verbally but
show signs of doubt or stress on their face.
Catching this mismatch early can help clinicians adjust their approach---for
instance during motivational interviewing~\cite{miller2012motivational}---or
flag attitudes linked to vaccine hesitancy~\cite{freeman2020vaccine} and
treatment refusal.

From a computational point of view, A/H is different from the ``Big~6'' basic
emotions.
Happiness or anger tend to show up clearly through a single channel: facial
action units, prosody, or word choice.
A/H, on the other hand, often produces \emph{contradictory} cross-modal
signals.
A person says ``I feel fine'' while their voice wavers.
Their words express agreement, but micro-expressions betray uncertainty.
Standard multimodal fusion methods are designed to find what the modalities
\emph{agree} on; they can actually suppress the disagreement signals that
define A/H.

The ABAW10 Ambivalence/Hesitancy Video Recognition
Challenge~\cite{kollias2024abaw10} provides a benchmark to study this
problem.
It uses the BAH dataset~\cite{gonzalez2026bah}: 1,427 clinical interview
videos from 300 participants, annotated at the video level with binary A/H
labels.
With only 778 training videos and expert-level annotations that are expensive
to obtain, the dataset is intentionally small---making regularisation and
pre-trained representations essential.

\noindent Our contributions are as follows:
\begin{enumerate}
  \item \textbf{Conflict-aware multimodal fusion.}
        We introduce explicit pairwise conflict features that act as
        \emph{bidirectional} cues: large cross-modal differences flag A/H,
        while small differences anchor the negative class.
        This corrects the systematic class bias of text-dominant approaches:
        our conflict-aware model improves F1-NoAH by +4.6 points over text
        alone and halves the class-performance gap---a crucial property for
        clinical deployment where false-positive A/H flags waste clinician
        time.
  \item \textbf{Text-guided late fusion.}
        A text-only auxiliary head is blended with the full multimodal output
        at inference, adding +4.1 Macro F1 by leveraging the strong
        linguistic signal while keeping multimodal evidence for ambiguous
        cases.
  \item \textbf{Efficient, reproducible baseline.}
        Our full pipeline trains in under 25 minutes on a single NVIDIA RTX
        4090, yet outperforms published multimodal baselines by more than 10
        points and achieves 0.715 Macro F1 on the official ABAW10
        leaderboard.
\end{enumerate}

\section{Related Work}

\paragraph{Affective Behaviour Analysis and the ABAW Challenges.}
The Affective Behaviour Analysis in-the-Wild (ABAW) workshop series has been a
principal driver of progress in multimodal affective computing, defining
benchmarks for facial action unit (AU) detection, valence-arousal estimation,
and discrete expression recognition~\cite{kollias2024abaw8}.
The eighth edition introduced the Ambivalence/Hesitancy (A/H) Recognition
Challenge, marking the first large-scale benchmark for this clinically motivated
task~\cite{kollias2024abaw8,kollias2024abaw10}.
The tenth edition (ABAW10)~\cite{kollias2024abaw10} continues this track with
the full BAH dataset~\cite{gonzalez2026bah}, providing participant-wise
train/val/test splits and an additional unlabelled hold-out for challenge
evaluation.

Several teams competed in ABAW-8.
Savchenko~\cite{savchenko2024abaw8} achieved the highest reported Macro F1 of
0.772 using a text-dominant approach: frame-level visual emotion features were
aggregated statistically and combined with transcript embeddings, with the
linguistic modality contributing the decisive gain.
This result established text as the primary carrier of A/H information.
Hallmen et al.~\cite{hallmen2025semantic} fused Wav2Vec~2.0 acoustic features
with BERT transcript embeddings and ViT visual features via a BiLSTM temporal
module, confirming that semantic content of speech is more discriminative than
acoustic prosody alone.
Collectively, these submissions converge on a modality
hierarchy---text $>$ audio $>$ visual---that motivates our text-guided late
fusion strategy.

\paragraph{BAH Dataset and A/H Recognition.}
Gonz\'alez et al.~\cite{gonzalez2026bah} introduced the BAH dataset and
established video-level multimodal baselines.
Their best multimodal fusion result---LFAN co-attention~\cite{zhang2023lfan}
over all three modalities---achieves 0.59 Macro F1 at video level; zero-shot
inference with a multimodal LLM~\cite{luo2024mllm,zhang2024gpt4v} and the full transcript reaches 0.63.
Annotator cue analysis reveals that facial--language inconsistency dominates
(42.6\% of annotated conflict segments), followed by language--body (26.6\%),
providing direct empirical motivation for our explicit conflict features.

\paragraph{Video Representation Learning.}
VideoMAE~\cite{tong2022videomae} extends masked autoencoders to video by
masking a high proportion of spatiotemporal tube patches and reconstructing
pixel values, producing representations that transfer well to action
recognition.
We adopt VideoMAE-Base (86M parameters) pre-trained on Kinetics-400, sampling
16 frames per clip.
VideoFocalNet~\cite{wasim2023videofocalnet} is an alternative video backbone
with strong results on the BAH video-level benchmark (Macro F1 0.566) using
facial crops; however, it provides visual features only and does not integrate
speech or transcript.

Temporal modelling is particularly relevant for A/H recognition: hesitation
cues accumulate over time~\cite{gonzalez2026bah}.
Prior work employs Temporal Convolutional Networks (TCN)~\cite{bai2018tcn}
and BiLSTM-based sequential models~\cite{hallmen2025semantic} for temporal
affective modelling.
In our architecture, multi-window inference---averaging predictions over five
randomly sampled 16-frame windows---provides implicit temporal coverage without
an explicit temporal decoder, a deliberate efficiency trade-off given the
778-sample training set.

\paragraph{Self-Supervised Speech Representations.}
HuBERT~\cite{hsu2021hubert} learns speech representations via offline k-means
cluster assignment as pseudo-labels, achieving strong performance on speech
tasks without supervision.
We use HuBERT-Base fine-tuned on LibriSpeech 960h (94M parameters) to encode
raw 16\,kHz mono audio, capturing prosodic cues---pauses, pitch variation,
stuttering---that annotators identify as primary A/H indicators in the BAH
codebook~\cite{gonzalez2026bah}.
Wav2Vec~2.0~\cite{baevski2020wav2vec} is an alternative initially explored but
replaced due to HuBERT's superior sensitivity to expressive clinical speech.

\paragraph{Emotion-Aware Language Models.}
GoEmotions~\cite{demszky2020goemotions} is a large-scale human-annotated
emotion dataset covering 27 fine-grained emotion categories including confusion,
doubt, and uncertainty.
\citeauthor{samlowe2022roberta}~\cite{samlowe2022roberta} fine-tuned
RoBERTa-Base~\cite{liu2019roberta} on GoEmotions, producing a model with
heightened sensitivity to hedging and hesitation language.
We adopt this model as our text encoder, encoding the full video transcript
rather than only the text aligned with the sampled visual window, reflecting
the video-level nature of A/H: verbal hesitation indicators may occur at any
point in the interview~\cite{gonzalez2026bah}.

\paragraph{Multimodal Fusion and Conflict Modelling.}
Multimodal fusion strategies are broadly categorised as early (feature-level),
late (decision-level), and hybrid~\cite{baltrusaitis2018multimodal}.
Our approach is hybrid: a conflict-augmented early fusion stage is followed by
a late blend with a unimodal text branch.
Element-wise absolute differences as conflict features are motivated by
disagreement-aware fusion in multimodal sentiment
analysis~\cite{poria2017context} and cross-modal mismatch modelling in
deception detection~\cite{perez2015verbal}.
Compound emotion recognition---where a mixture of basic emotions is expressed
simultaneously---shares the cross-modal conflict structure of A/H;
C-EXPR-DB~\cite{kollias2023cexpr} demonstrates that modality representations
must capture contradictions rather than consensus, directly supporting our
conflict feature design.

\paragraph{Attention Pooling.}
Reducing variable-length encoder outputs to fixed-size representations is
standard in multimodal pipelines.
Learnable soft-attention pooling~\cite{bahdanau2015attention} assigns
importance weights to each token via a learned query vector, enabling the
model to upweight emotionally salient moments---vocal hesitation pauses,
uncertain facial expressions---while downweighting uninformative tokens.
This approach is well-established in audio-visual emotion
recognition~\cite{praveen2023crossattention} and has been applied effectively
to speech-based affective tasks.
We employ modality-specific attention pooling for each encoder output, with
padding masking for variable-length audio.

\paragraph{Learning from Small Affective Datasets.}
The BAH training set contains 778 labelled videos---a scale typical of
expert-annotated clinical data.
Freezing pre-trained encoder parameters and fine-tuning only task-specific
heads is a well-established approach for small-data affective computing,
preventing catastrophic forgetting while reducing the effective parameter count.
Label smoothing~\cite{szegedy2016rethinking} regularises the training signal
by softening hard targets, reducing overconfidence.
Threshold calibration on the validation set---selecting the operating point
that maximises Macro F1 rather than defaulting to 0.5---is standard practice
in binary affective challenge evaluation with class
imbalance~\cite{savchenko2024abaw8}.
Checkpoint ensembling provides additional regularisation by averaging
predictions from models trained under different regularisation settings.

\section{Method}

\subsection{Problem Formulation}

Given a video $\mathcal{V}$ with aligned audio $\mathcal{A}$ and a
video-level speech transcript $\mathcal{T}$, the goal is to predict a binary
label $y \in \{0, 1\}$ indicating whether Ambivalence/Hesitancy is present.

Following the challenge protocol~\cite{gonzalez2026bah}, the primary metric is
\textbf{Macro F1} (AVGF1): the unweighted average of F1 for the positive
class (A/H present, denoted F1-AH) and F1 for the negative class (A/H absent,
denoted F1-NoAH).

\subsection{Multimodal Encoders}

We use three modality-specific pre-trained encoders.
Table~\ref{tab:encoders} summarises our choices.

\begin{table}[h]
  \centering
  \caption{Encoder backbones.
  All three output 768-dimensional embeddings and were selected for strong
  pre-trained representations while keeping the model footprint moderate.}
  \label{tab:encoders}
  \small
  \begin{tabular}{llcc}
    \toprule
    Modality & Backbone & Params & Dim \\
    \midrule
    Video & VideoMAE-Base~\cite{tong2022videomae}    & 86M  & 768 \\
    Audio & HuBERT-Base-LS960~\cite{hsu2021hubert} & 94M  & 768 \\
    Text  & RoBERTa-GoEmotions~\cite{samlowe2022roberta,demszky2020goemotions} & 125M & 768 \\
    \bottomrule
  \end{tabular}
\end{table}

\noindent
\textbf{Video.}
We sample $T{=}16$ contiguous frames from each video.
During training, the starting position is chosen randomly (temporal
augmentation).
At inference, we sample five independent windows and average their predictions
(Section~\ref{sec:training}).
All frames use the cropped-aligned faces provided by the BAH dataset and are
normalised with ImageNet statistics at $224{\times}224$ resolution.
VideoMAE~\cite{he2022masked} processes each sequence as pixel-value patches
and outputs a token sequence of shape $(B, L_v, 768)$.

\noindent
\textbf{Audio.}
Raw 16\,kHz mono audio is passed directly to HuBERT, producing a frame-level
sequence $(B, L_a, 768)$.
Variable-length audio is zero-padded within each batch; the padding mask is
carried forward to the attention pooling step so that padded positions do not
influence the representation.

\noindent
\textbf{Text.}
The BAH dataset provides automatically generated transcripts produced using
Whisper ASR~\cite{radford2023whisper}.
The \emph{full} video-level transcript is tokenised and encoded by
RoBERTa-GoEmotions, yielding $(B, L_t, 768)$.
We deliberately use the complete transcript rather than only the few seconds
that correspond to the sampled video window, because A/H is a
\emph{video-level} property and hesitation cues can appear anywhere in the
interview.
The 27 fine-grained emotion categories of GoEmotions (e.g.\ confusion,
nervousness, doubt) match the hedging and uncertainty vocabulary that
characterises A/H, making this encoder more suitable than BERT trained only on
generic text.

\subsection{Attention Pooling}

Each encoder produces a variable-length sequence $(B, L, D)$ that we reduce to
a single vector $(B, D)$ via learned soft-attention pooling:
\begin{equation}
  \alpha_i = \frac{\exp(\mathbf{w}^\top \mathbf{x}_i)}
                  {\sum_j \exp(\mathbf{w}^\top \mathbf{x}_j)},
  \quad
  \mathbf{e} = \sum_i \alpha_i \mathbf{x}_i,
\end{equation}
where $\mathbf{w} \in \mathbb{R}^D$ is a learnable query vector and padding
positions are masked to $-\infty$ before the softmax.
A modality-specific linear projection maps each encoder's output to the shared
dimension $D{=}768$ before pooling.
Attention pooling lets the model place more weight on the most discriminative
time steps---a hesitation marker or an emotionally loaded word---rather than
averaging all tokens equally.

\subsection{Conflict-Aware Fusion}

The central idea behind our fusion is simple: A/H is, by definition, a state
of internal conflict.
The person simultaneously holds opposing attitudes.
We hypothesise that this internal tension leaks into observable
\emph{cross-modal disagreement}.
For example, if the transcript sounds positive but the face shows stress, that
mismatch \emph{is} the signal.

To capture this, we compute pairwise absolute differences between all modality
embeddings:
\begin{align}
  \mathbf{c}_{va} &= |\mathbf{v} - \mathbf{a}|, \\
  \mathbf{c}_{vt} &= |\mathbf{v} - \mathbf{t}|, \\
  \mathbf{c}_{at} &= |\mathbf{a} - \mathbf{t}|,
\end{align}
where $\mathbf{v}, \mathbf{a}, \mathbf{t} \in \mathbb{R}^D$ are the pooled
video, audio, and text embeddings.
The absolute difference is large along dimensions where two modalities
diverge, and the downstream classifier learns which of these disagreements
matter for A/H prediction.
This choice is motivated by the BAH annotator analysis~\cite{gonzalez2026bah},
which shows that face--language inconsistency accounts for 42.6\% of annotator
cues.
Absolute difference has also proven effective for cross-modal inconsistency in
sentiment analysis~\cite{poria2017context}.

Note that the modality-specific projections are learned end-to-end and are not
pre-aligned through contrastive objectives (e.g.\ CLIP-style pre-training).
The alignment emerges entirely from the classification signal.

The full fusion representation concatenates modality and conflict features:
\begin{equation}
  \mathbf{f} = [\mathbf{v};\, \mathbf{a};\, \mathbf{t};\,
               \mathbf{c}_{va};\, \mathbf{c}_{vt};\, \mathbf{c}_{at}]
  \in \mathbb{R}^{6D}.
\end{equation}
A two-layer feed-forward network (LayerNorm, Linear$\to$GELU$\to$Dropout,
hidden width $2{\times}6D$) processes this vector,
followed by a classification MLP head:
LayerNorm $\to$ Linear$(6D, 512)$ $\to$ GELU $\to$ Dropout$(0.3)$ $\to$
Linear$(512, 1)$,
producing the full-fusion logit $\ell_{\mathrm{full}} \in \mathbb{R}$.

\subsection{Text-Guided Late Fusion}
\label{sec:latfusion}

Text carries the strongest discriminative signal for A/H---a finding reported
by Savchenko~\cite{savchenko2024abaw8} and confirmed by our own modality
ablation (Table~\ref{tab:ablation_modality}).
Rather than ignoring this and hoping the multimodal path figures it out, we
make it explicit with a parallel \emph{text-only auxiliary head}~\cite{liang2024auxiliary}:
\begin{equation}
  \ell_{\mathrm{text}} = \mathrm{MLP}_{\mathrm{text}}(\mathbf{t}),
\end{equation}
identical in architecture to the full-fusion head but operating only on
$\mathbf{t} \in \mathbb{R}^D$.

\noindent\textbf{Why does this help?}
On a small dataset like BAH (778 training videos), the multimodal path can
overfit to noisy visual or acoustic patterns.
The text-only head acts as a regulariser: it forces the shared text encoder to
produce embeddings that are independently discriminative, not just useful when
combined with other modalities.

\noindent\textbf{Training.}
Both heads are trained jointly with balanced loss weights:
\begin{equation}
  \tilde{y} = y(1-\varepsilon) + 0.5\varepsilon,
\end{equation}
\begin{equation}
  \mathcal{L} = (1-w)\,\mathrm{BCE}(\ell_{\mathrm{full}}, \tilde{y})
              + w\,\mathrm{BCE}(\ell_{\mathrm{text}}, \tilde{y}),
\end{equation}
where $w{=}0.5$ and $\varepsilon$ is a label-smoothing parameter
($\varepsilon{=}0.1$ or $0$ depending on the checkpoint configuration;
see Section~\ref{sec:impl}).

\noindent\textbf{Inference.}
At test time the two branches are blended:
\begin{equation}
  p = \alpha \cdot \sigma(\ell_{\mathrm{text}})
    + (1-\alpha) \cdot \sigma(\ell_{\mathrm{full}}),
  \label{eq:blend}
\end{equation}
where $\sigma$ is the sigmoid function and $\alpha{=}0.6$, tuned on the
validation set.
The weight $\alpha > 0.5$ reflects the text branch's stronger predictive
power while still incorporating multimodal evidence.

\subsection{Architecture Overview}

\begin{figure}[t]
  \centering
  \includegraphics[width=\columnwidth]{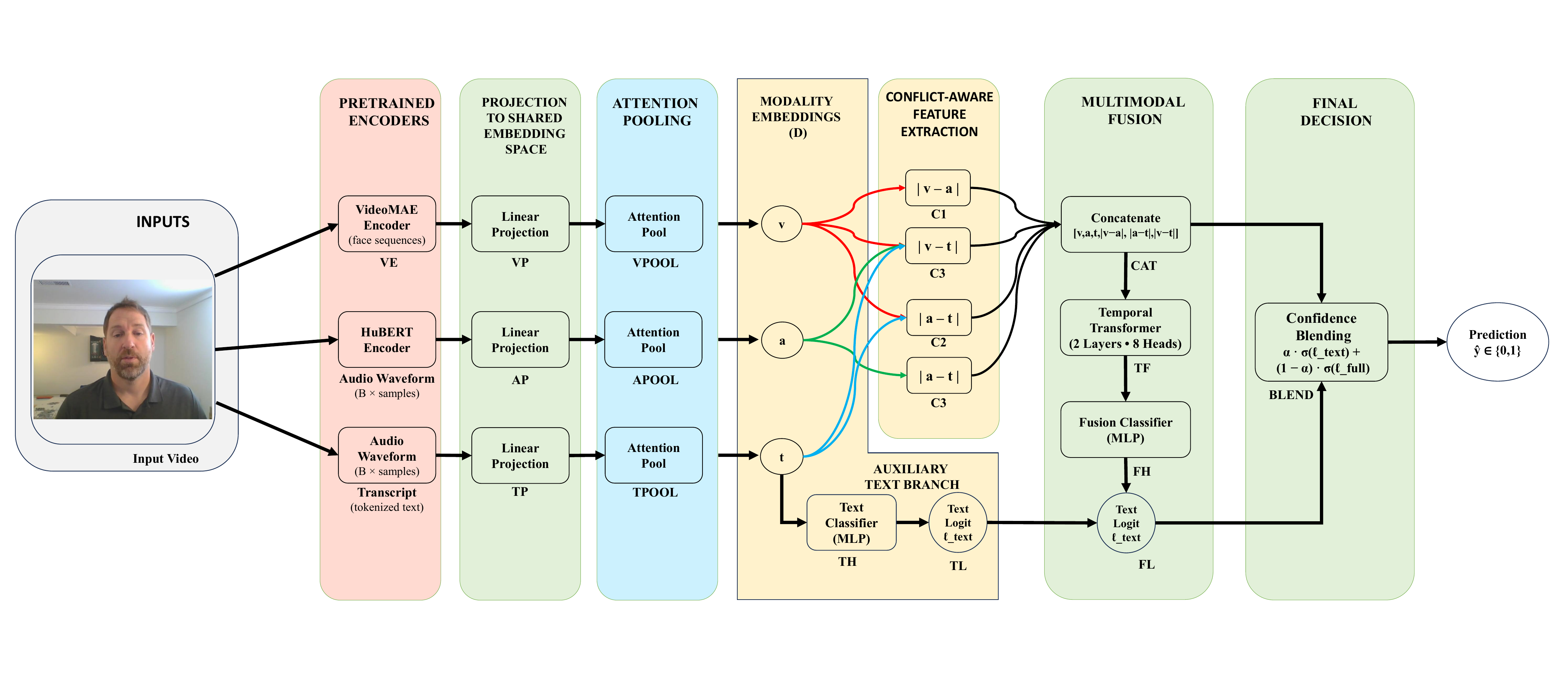}
  \caption{
    \textbf{ConflictAwareAH architecture.}
    Video frames, audio waveform, and transcript pass through pre-trained
    encoders and attention pooling to produce embeddings
    $\mathbf{v}$, $\mathbf{a}$, $\mathbf{t}$.
    Three pairwise conflict vectors are concatenated with the modality
    embeddings and fed to a two-layer FFN, producing the full-fusion logit
    $\ell_{\mathrm{full}}$.
    A parallel text-only head produces $\ell_{\mathrm{text}}$.
    At inference the two are blended via Eq.~\ref{eq:blend}.
  }
  \label{fig:arch}
\end{figure}

Figure~\ref{fig:arch} gives a high-level view of the full pipeline.

\subsection{Training Protocol}
\label{sec:training}

\begin{figure}[t]
  \centering
  \includegraphics[width=\columnwidth]{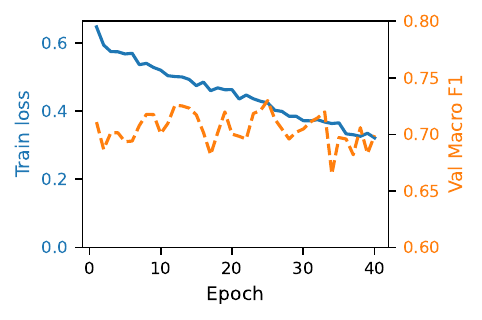}
  \caption{Train loss and validation Macro F1 over epochs (Run A).}
  \label{fig:training}
\end{figure}

\noindent\textbf{Optimiser.}
AdamW ($\beta_1{=}0.9$, $\beta_2{=}0.999$, weight decay $10^{-2}$) with
learning rate $3{\times}10^{-5}$ and cosine annealing to $3{\times}10^{-7}$.
Mixed precision (fp16) is used throughout.

\noindent\textbf{Batch size.}
We use mini-batches of 4, accumulated over 4 gradient steps for an effective
batch size of 16.

\noindent\textbf{Encoder freezing.}
Given only 778 training samples, we freeze all encoder weights by default.
The top-2 Transformer layers of each encoder are optionally unfrozen at a
$10\times$ lower learning rate ($3{\times}10^{-6}$) for gentle adaptation.

\noindent\textbf{Class weighting.}
We re-weight BCEWithLogitsLoss by the inverse positive-class frequency
($\text{pos\_weight} \approx 0.96$).

\noindent\textbf{Early stopping.}
Training stops when Macro F1 on the validation set does not improve for 15
consecutive epochs (max 60 epochs).
Figure~\ref{fig:training} shows typical train loss and validation Macro F1
curves; train/val divergence after epoch~17--19 indicates overfitting.

\noindent\textbf{Threshold tuning.}
After each epoch we sweep thresholds in $[0.25, 0.75]$ at 0.01 resolution
on the blended validation probabilities.
The best threshold from the checkpoint with peak Macro F1 is used at test
time.

\noindent\textbf{Multi-window inference.}
At test time, instead of a single 16-frame window, we can sample $N$
random contiguous windows per video and average the resulting probabilities.
This gives broader visual coverage over the whole interview, better matching
the temporal scope of the full transcript.
We use $N{=}5$ in our submitted results.

\section{Experiments}

\subsection{Dataset}

All experiments use the BAH dataset provided for the ABAW10
challenge~\cite{gonzalez2026bah,kollias2024abaw10}.
It contains 1,427 videos (10.6 hours, $\sim$27\,s per video on average) from
300 Canadian participants.
Each video has video-level and frame-level A/H annotations, a speech-to-text
transcript, and participant metadata.
Splits are made at the participant level so that no person appears in both
training and evaluation.
Table~\ref{tab:dataset} shows the statistics.

\begin{table}[h]
  \centering
  \caption{BAH dataset split statistics for the ABAW10 A/H challenge.}
  \label{tab:dataset}
  \small
  \begin{tabular}{lrrr}
    \toprule
    Split & Videos & A/H (+) & No A/H ($-$) \\
    \midrule
    Train        & 778  & 385 (49\%) & 393 (51\%) \\
    Val          & 124  & 75 (60\%)  & 49 (40\%)  \\
    Test         & 525  & 318 (61\%) & 207 (39\%) \\
    Test (unlabeled) & 161 & --- & --- \\
    \bottomrule
  \end{tabular}
\end{table}

\subsection{Implementation Details}
\label{sec:impl}

All experiments run on a single NVIDIA RTX 4090 GPU (24\,GB).
Pre-trained weights come from the HuggingFace Model Hub.
Training one epoch takes roughly 50 seconds; a typical run converges in 20--25
minutes.

For our final submission we use an \emph{ensemble of two checkpoints}, each
trained with the same architecture but different regularisation settings:
\begin{itemize}
  \item \textbf{Checkpoint A:} top-2 encoder layers unfrozen, no label
        smoothing ($\varepsilon{=}0$).
  \item \textbf{Checkpoint B:} all encoder layers fully frozen, label
        smoothing with $\varepsilon{=}0.1$.
\end{itemize}
The two models produce complementary predictions: Checkpoint A benefits from
task-specific encoder adaptation while Checkpoint B is more stable and less
prone to overfitting.
At test time, we average the sigmoid probabilities from both checkpoints
(each using 5-window inference) and then apply the best validation threshold.
The code is implemented in PyTorch~\cite{paszke2019pytorch} with HuggingFace
Transformers~\cite{wolf2020transformers}.

\subsection{Main Results}

Table~\ref{tab:main} compares ConflictAwareAH against published baselines.

\begin{table}[h]
  \centering
  \caption{
    Comparison on the ABAW10 A/H test split (525 videos).
    BAH baselines are from Gonz\'alez et al.~\cite{gonzalez2026bah}.
  }
  \label{tab:main}
  \begin{tabular}{lc}
    \toprule
    Method & Macro F1 \\
    \midrule
    BAH: Zero-shot M-LLM (vision only)~\cite{gonzalez2026bah} & 0.283 \\
    BAH: Video-FocalNet base~\cite{gonzalez2026bah} & 0.566 \\
    BAH: LFAN (V+A+T, co-attention)~\cite{gonzalez2026bah} & 0.593 \\
    BAH: Zero-shot M-LLM + transcript~\cite{gonzalez2026bah} & 0.634 \\
    \midrule
    Ours --- single model (1 or 5 windows) & 0.690--0.692 \\
    \textbf{Ours --- ensemble (2 ckpts $\times$ 5 win.)} & \textbf{0.694} \\
    \bottomrule
  \end{tabular}
\end{table}

A single ConflictAwareAH model already outperforms the strongest BAH
baseline (M-LLM + transcript, 0.634) by +5.6 points.
The two-checkpoint ensemble pushes this to 0.694.
Importantly, training a single model takes under 25 minutes on one RTX 4090
GPU---no multi-node training, no large-scale fine-tuning.

On the private \textbf{test\_unlabeled} split (161 videos, labels withheld),
our submitted ensemble obtained an official Macro F1 of \textbf{0.715} in the
ABAW10 challenge evaluation.
The improvement from 0.694 (labelled test) to 0.715 (private test) suggests the
model generalises robustly to unseen participants.

\subsection{Ablation Studies}

We now examine each component of our design through systematic ablations.
Unless stated otherwise, experiments use the full model with $\alpha{=}0.6$
(text blend), top-2 layer unfreezing, no label smoothing, and single-window
inference on the test split.

\paragraph{How much does the text blend help?}
Table~\ref{tab:ablation_components} answers this question.
Setting $\alpha{=}0.0$ (full-fusion only, no text branch at inference) gives
0.649 Macro F1.
Blending with $\alpha{=}0.6$ raises this to 0.690:
a \textbf{+4.1 point gain} from a single hyperparameter.
In fact, using the text head alone ($\alpha{=}1.0$) achieves 0.700, the best
single-branch score.
The text-guided blend is clearly the primary driver of our performance.

\paragraph{Do conflict features help?}
Removing the three conflict vectors and feeding only
$[\mathbf{v};\,\mathbf{a};\,\mathbf{t}]$ to the fusion network yields 0.681.
Our full model (0.690) outperforms this by \textbf{+0.9 points}, indicating
that the explicit conflict features contribute to the model's discriminative
ability.
While the gain is modest, it is consistent with the hypothesis that
cross-modal disagreement carries a meaningful signal for A/H.

\paragraph{Full transcript vs.\ window-aligned text.}
Restricting the text to only the seconds matching the video window gives 0.698,
comparable to using the full transcript (0.690).
We use the full transcript by default since A/H is a video-level label and
hesitation markers can occur at any point in the interview.

\begin{table}[h]
  \centering
  \caption{
    Component ablation on the test split.
  }
  \label{tab:ablation_components}
  \small
  \begin{tabular}{lc}
    \toprule
    Configuration & Macro F1 \\
    \midrule
    No conflict features (v+a+t only) & 0.681 \\
    Window-aligned transcript only & 0.698 \\
    \midrule
    Text blend $\alpha = 0.0$ (full-fusion only) & 0.649 \\
    Text blend $\alpha = 0.5$                    & 0.689 \\
    \textbf{Text blend $\alpha = 0.6$ (ours)}   & \textbf{0.690} \\
    Text blend $\alpha = 1.0$ (text-only at inference) & 0.700 \\
    \bottomrule
  \end{tabular}
\end{table}

\paragraph{Which modalities matter most?}
Table~\ref{tab:ablation_modality} breaks this down on the validation split.
Text alone reaches 0.737 Macro F1---far above video (0.534) or audio (0.523).
Combining all three modalities (0.746) yields a modest gain over text only
(+0.9 points).
Adding video or audio to text does not hurt, but the improvement is small,
reflecting the strong dominance of the linguistic channel for this task.
However, multimodal cues do improve robustness when linguistic signals are
ambiguous---for instance during pauses, filled hesitations, or when the
transcript is noisy.

\begin{table}[h]
  \centering
  \caption{
    Modality ablation (validation split, 124 videos).
  }
  \label{tab:ablation_modality}
  \small
  \begin{tabular}{lccc}
    \toprule
    Modality & Macro F1 & F1-AH & F1-NoAH \\
    \midrule
    Video only     & 0.534 & 0.560 & 0.508 \\
    Audio only     & 0.523 & 0.645 & 0.400 \\
    Text only      & 0.737 & 0.804 & 0.670 \\
    \midrule
    Video + Audio  & 0.527 & 0.573 & 0.481 \\
    Video + Text   & 0.734 & 0.783 & 0.685 \\
    Audio + Text   & 0.736 & 0.789 & 0.683 \\
    \midrule
    \textbf{All three (ours)} & \textbf{0.746} & 0.777 & 0.716 \\
    \bottomrule
  \end{tabular}
\end{table}

\paragraph{Inference windows and ensembling.}
Table~\ref{tab:ablation_training} shows how multi-window inference and
checkpoint ensembling affect the test score.
Going from 1 window to 5 windows has marginal effect on its own, but the
two-checkpoint ensemble consistently provides the best result (0.694), likely
because the two checkpoints are trained with different regularisation and
therefore make different errors.

\begin{table}[h]
  \centering
  \caption{
    Inference settings on the test split.
  }
  \label{tab:ablation_training}
  \small
  \begin{tabular}{lc}
    \toprule
    Setting & Macro F1 \\
    \midrule
    Inference: 1 window   & 0.692 \\
    Inference: 3 windows  & 0.690 \\
    Inference: 5 windows (ours) & 0.690 \\
    \midrule
    Single checkpoint (5 windows) & 0.690 \\
    \textbf{Ensemble of 2 checkpoints (5 win.)} & \textbf{0.694} \\
    \bottomrule
  \end{tabular}
\end{table}

\subsection{Analysis and Discussion}

\paragraph{Text dominates---but is biased.}
The ablation results tell a consistent story: text is the most informative
individual modality for A/H recognition.
Using only the text branch at inference ($\alpha{=}1.0$) reaches 0.700 Macro
F1, higher than full-fusion only ($\alpha{=}0.0$, 0.649).
However, a closer look at Table~\ref{tab:ablation_modality} reveals a
systematic weakness: text-only achieves high F1-AH (0.804) but substantially
lower F1-NoAH (0.670), yielding a class-performance gap of 0.134.
The model over-detects A/H because hedging and uncertainty language is common
in clinical interviews even when no genuine ambivalence is present.
In contrast, our full conflict-aware model narrows this gap to 0.061
(F1-AH~0.777, F1-NoAH~0.716),
improving F1-NoAH by \textbf{+4.6 points} while sacrificing only 2.7 points on
F1-AH.
This rebalancing is clinically important: in deployment, false-positive A/H
flags waste clinician time, so a model that reliably confirms the
\emph{absence} of ambivalence is at least as valuable as one that detects its
presence.

\paragraph{Conflict features as a bidirectional signal.}
The class-bias correction above is driven by the conflict features.
When the modality embeddings \emph{agree}---$|\mathbf{v}-\mathbf{t}|$ is
small---the conflict vector provides an explicit consistency cue that anchors
the negative class.
A text-only approach has no mechanism to verify whether verbal content
matches facial and prosodic behaviour; it can detect hesitant \emph{language}
but cannot confirm behavioural \emph{consistency}.
Conversely, when the transcript expresses certainty (e.g.\ ``I'm totally fine
with it'') but the face shows micro-expressions of stress, the conflict vector
is large along the relevant dimensions and the classifier correctly flags A/H.
Thus, the conflict features act as a bidirectional signal: disagreement flags
A/H, agreement confirms No~A/H.
On the aggregate metric, this appears as a modest +0.9 Macro F1
(Table~\ref{tab:ablation_components}); the per-class breakdown reveals that
the real contribution is a substantially better-calibrated model.

\paragraph{Text-guided late fusion complements conflict features.}
Text-guided late fusion adds +4.1 Macro F1 by allowing the system to rely on
the text head when its linguistic signal is confident, while still having
access to multimodal and conflict-based evidence for harder cases.
The two mechanisms address different failure modes: conflict features correct
class bias, while text-guided blending boosts overall discriminability.

\paragraph{Robustness beyond text.}
Relying exclusively on text introduces additional risks in clinical
deployment.
Automatic transcripts are generated by Whisper ASR~\cite{radford2023whisper},
which can produce errors on accented, mumbled, or overlapping clinical
speech.
Silent or minimally verbal patients---especially those exhibiting
non-verbal A/H cues---would yield uninformative transcripts.
Moreover, clinicians primarily assess A/H through visual and prosodic cues
(the BAH annotator analysis attributes 42.6\% of cues to facial--language
inconsistency~\cite{gonzalez2026bah}).
A multimodal conflict-aware system aligns more closely with clinical practice
than a text-only approach.

\paragraph{Overfitting is the main challenge.}
With 778 training samples and potentially 60M trainable parameters (when
encoder layers are unfrozen), the train--validation loss curves diverge after
epoch 17--19.
Freezing all encoder layers and applying label smoothing produces comparable
peak performance with faster convergence and smaller generalisation gaps,
making it the safer choice at this data scale.

\paragraph{Limitations.}
First, threshold tuning on the validation set may not transfer perfectly if
the test distribution shifts significantly; we report both splits for
transparency.
Second, the 16-frame window captures only a small portion of each video;
multi-window inference alleviates this but does not fully solve temporal
modelling.
Third, the modality embeddings are not pre-aligned through contrastive
objectives; the absolute-difference conflict features therefore capture
geometric rather than semantic distance, and richer conflict representations
(e.g.\ cross-attention between modalities or projected differences in a
shared space) are a promising direction for future work.

\section{Conclusion}

We presented ConflictAwareAH, a multimodal framework for recognising
Ambivalence and Hesitancy in clinical interview videos.
The key insight is that while text is the most discriminative individual
modality for A/H, it is also systematically biased: text-only models
over-detect A/H while struggling to confirm its absence.

Our main contribution, \emph{conflict-aware fusion}, addresses this through
explicit pairwise cross-modal difference features.
These features serve as bidirectional cues---large differences flag A/H
while small differences anchor the negative class---improving F1-NoAH by
+4.6 points and halving the class-performance gap compared to text alone.
A complementary \emph{text-guided late fusion} strategy adds +4.1 Macro F1
by blending a text-only auxiliary head with the full multimodal output.
Our best result (0.694 Macro F1 on the labelled test split, 0.715 on the
ABAW10 private leaderboard) comes from ensembling two checkpoints trained with
complementary regularisation and averaging predictions over multiple inference
windows.

The system outperforms all published BAH multimodal baselines (best: 0.593)
and the zero-shot M-LLM with transcript (0.634), and it trains in under 25
minutes on a single GPU.

Three directions for future investigation stand out:
(1)~alignment-aware conflict representations through contrastive pre-training
or cross-attention, which may amplify the bidirectional signal;
(2)~temporal modelling across multiple video segments, where a Transformer
attends over $N$ independent clip representations instead of averaging them;
and
(3)~richer regularisation strategies (e.g.\ mixup, stochastic depth, or
parameter-efficient fine-tuning) to push performance further on
datasets of this scale.

{\small
\bibliographystyle{ieeenat_fullname}
\bibliography{references}
}

\end{document}